\documentclass[10pt,letterpaper]{article}
\usepackage[pagenumbers]{cvpr} 
\usepackage{multirow}
\usepackage{graphicx}
\usepackage{booktabs}
\usepackage{bibentry}
\usepackage{overpic}
\usepackage[utf8]{inputenc}
\usepackage{amsmath}
\usepackage{amsfonts}
\usepackage{amssymb}
\usepackage{graphicx}
\usepackage{booktabs}
\usepackage{array}
\usepackage{multirow}
\usepackage{adjustbox}
\usepackage{xcolor}

\usepackage[T1]{fontenc}

\usepackage{colortbl} 
\definecolor{cvprblue}{rgb}{0.21,0.49,0.74}
\usepackage[pagebackref,breaklinks,colorlinks,allcolors=cvprblue]{hyperref}

\def\paperID{14810} 
\def\confName{CVPR}
\def\confYear{2026}

\title{DCoAR: Deep Concept Injection into Unified Autoregressive Models for \\ Personalized Text-to-Image Generation}

\author{
    Fangtai Wu\textsuperscript{\rm 1}\footnotemark[1] \and
    Mushui Liu\textsuperscript{\rm 1,2}\footnotemark[1] \and
    Weijie He\textsuperscript{1} \and
    Zhao Wang\textsuperscript{1} \and
    Yunlong Yu\textsuperscript{\rm 1}\footnotemark[2]
    \vspace{0.5em} \\ 
    \textsuperscript{\rm 1}Zhejiang University \qquad 
    \textsuperscript{\rm 2}Alibaba Group
}
\begin{document}
\twocolumn[{%
\renewcommand\twocolumn[1][]{#1}%
\maketitle
\vspace{-1em}
\includegraphics[width=\linewidth]{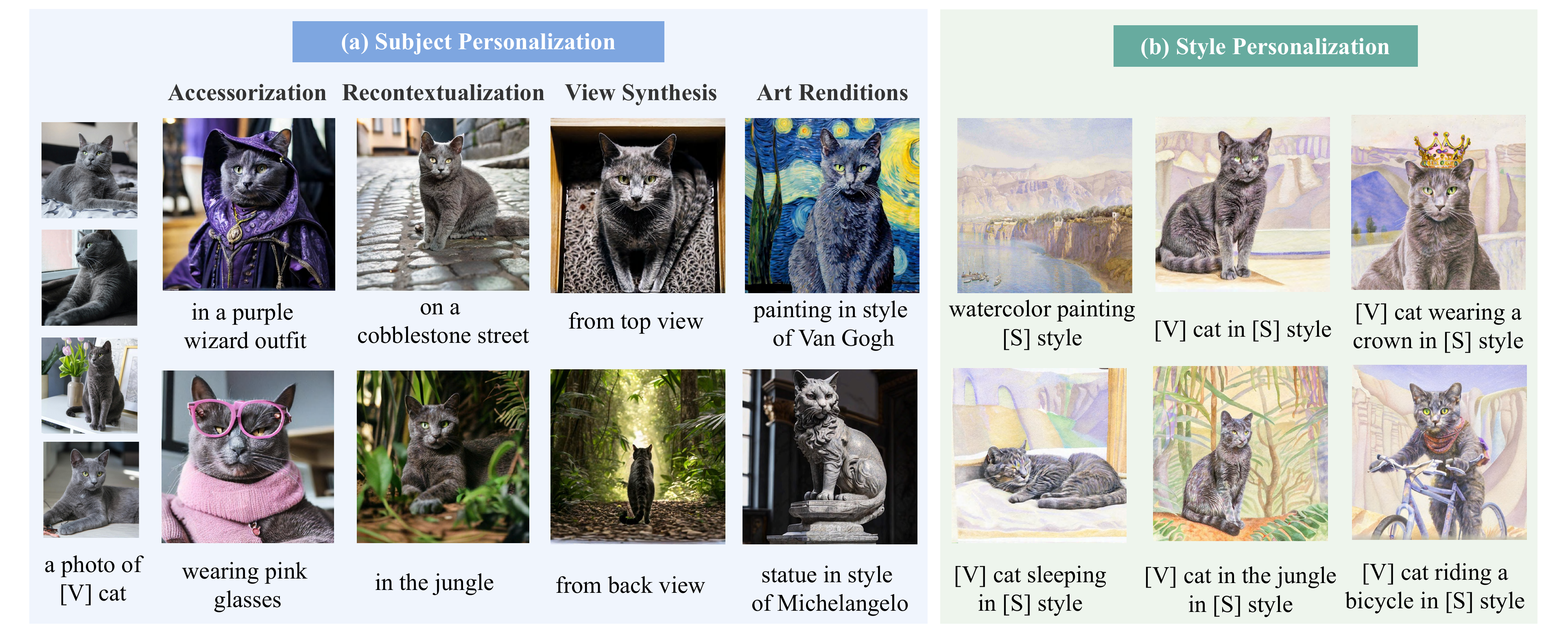}
\captionof{figure}{Visualization of samples generated by our proposed DCoAR, showcasing subject personalization (left), where a few reference images are provided, and style personalization (right), where the subject is rendered in user-specified artistic styles without additional training. \vspace{2em}
}
\label{fig:teaser}
}]
\maketitle

\begingroup
    \renewcommand{\thefootnote}{\fnsymbol{footnote}}
    \footnotetext[1]{Equal contribution.} 
    \footnotetext[2]{Corresponding authors.}
\endgroup

\begin{abstract}
The unified autoregressive (AR) model excels at multimodal understanding and generation. However, its full potential in the domain of customized image generation has yet to be fully realized.
Existing customization approaches for unified AR models face a fundamental dilemma: adaptation-based methods suffer from overfitting and scalability bottlenecks, while concept-injection paradigms are constrained by a shallow injection strategy that leads to poor visual fidelity and impaired re-contextualization.
To address this, we propose DCoAR, a novel deep concept injection framework that maintains a completely frozen pre-trained model. DCoAR deeply integrates new concepts through a Layer-wise Multimodal Context Learning (LMCL) strategy, which is stabilized by a multi-faceted regularization scheme: a Dual Prior Preservation (DPP) loss to mitigate semantic drift and a Context-Aware Self-Regularization (CASR) loss to enhance re-contextualization. The framework also enables training-free subject customization in user-provided styles.
Experiments demonstrate that DCoAR significantly outperforms previous injection-based methods and achieves performance competitive with adaptation-based approaches while requiring substantially fewer trainable parameters.

\end{abstract}    
\section{Introduction}
\label{sec:intro}
Unified multimodal autoregressive (AR) models \cite{team2024chameleon,lumina_mgpt,mmr1,stylear,editar,acdit}, have established a powerful paradigm for processing interleaved text and image data, demonstrating remarkable success across a range of generative and understanding tasks. Despite their success, the application of these models to personalized image generation has received comparatively less attention, leaving their full potential in this specific task open for exploration.
Personalized image generation\cite{chen2023subject,ruiz2023dreambooth,li2023blip,shin2025large,liu2024llm4genleveragingsemanticrepresentation,mosaicmultisubjectpersonalizedgeneration,TFcustom,she2025customvideox3dreferenceattention} aims to generate images of a specific subject or concept in new scenes, given only a few reference images.
This capability has found applications in virtual try-on \cite{ye2025slot,zhang2024two}, personalized content creation \cite{liu2024make}, and advertising design \cite{chen2023subject}.

Personalization for unified AR models remains fundamentally challenging, and existing approaches fall short in ways that are increasingly incompatible with real-world demands. On the one hand, fine-tuning–based methods, inspired by DreamBooth \cite{ruiz2023dreambooth,chen2023disenbooth,han2023svdiff,customdiff,OmniControl,EasyControl} and implemented through PEFT techniques such as LoRA \cite{Prefixtuning,coop,hu2022lora,personalAR,wu2025proxytuning}, attempt to rewrite model weights for each new concept. Yet AR models are far more brittle than diffusion models: adapting millions of parameters using only 3–5 images almost inevitably causes overfitting, catastrophic forgetting, and distortion of pre-trained priors. Worse still, maintaining a separate set of weights for every concept leads to unbounded storage costs, making this strategy fundamentally non-scalable for consumer-facing or large-scale personalization systems. On the other hand, token-injection approaches \cite{UniCTokens,yoChameleon} promise scalability by keeping the backbone frozen, but they suffer from a deeper structural flaw. Injecting concept information only at the input layer forces the signal to pass through dozens of transformer blocks without reinforcement. This shallow injection bottleneck results in severe semantic attenuation, where fine-grained identity cues vanish as depth increases, and the model fails to bind the concept to complex prompts. Consequently, these methods struggle with visual fidelity, contextual adaptability, and semantic consistency, falling short of fine-tuning counterparts.

To overcome the constraints of shallow concept injection, we introduce \textbf{D}eep \textbf{Co}ncept Injection for unified \textbf{AR} models (DCoAR), an efficient and extensible framework for customized image generation. The core of our DCoAR framework is the \textbf{Layer-wise Multimodal Context Learning (LMCL) strategy} that encodes subject-specific information into a compact set of learnable tokens and injects them across multiple transformer layers, enabling deeper semantic alignment and high-fidelity feature propagation \textit{without modifying the frozen backbone}. To ensure the robustness and quality of this deep injection process, DCoAR incorporates a multi-faceted regularization scheme consisting of two components. First, a \textbf{Dual Prior Preservation (DPP)} loss constrains the model against semantic drift via constraining the concept-conditioned features to stay close to the model’s native feature distribution while penalizing deviations from its latent feature gradients, ensuring that personalization does not compromise the inherent generative priors of the base AR model. Second, a \textbf{Context-Aware Self-Regularization (CASR)} enhances re-contextualization by preventing the concept tokens from over-specializing to the limited training prompts. CASR enforces cross-context consistency by aligning the model’s internal representations under both original and perturbed contexts, encouraging the tokens to stay flexible rather than memorizing appearance details. It further penalizes representation collapse across layers, ensuring that the injected concept signals remain discriminative and generalizable during deep propagation.

Our main contributions are summarized as follows:
\begin{itemize}
    \item We propose a deep concept injection framework for unified AR models. Unlike prior methods constrained by shallow, early-layer conditioning, DCoAR introduces a Layer-wise Multimodal Context Learning (LMCL) strategy that injects concept signals throughout the transformer stack, enabling persistent feature reinforcement and richer cross-modal interaction. This design achieves robust concept propagation and high-fidelity personalization without modifying a single parameter of the frozen backbone.    
    \item We introduce a training-free subject–style composition strategy. By directly concatenating the learned context tokens for a subject and a style, DCoAR enables arbitrary, plug-and-play combinations at inference time, achieving high-quality, context-aware stylization without any additional optimization or fine-tuning. Some generated samples are shown in \cref{fig:teaser}.
    \item Experimental results show that DCoAR sets a new state of the art for unified AR model personalization. Furthermore, DCoAR delivers strong gains in subject fidelity, instruction alignment, and parameter efficiency, requiring fewer than 0.1M trainable parameters. For stylization, our training-free pipeline produces high-fidelity results that rival modern LoRA-based methods without any additional training.

\end{itemize}

\section{Related Works}
\subsection{Autoregressive Image Generation}
In recent years, autoregressive (AR) image generation has rapidly evolved from pixel-level sequential modeling to token-level generation and further to multi-modal unified architectures. Early approaches \cite{pixelcnn,pixelrnn} operated directly in the pixel space, which imposed heavy computational demands for generating high-resolution images. The VQ-VAE family \cite{dalle,vqvae} subsequently discretizes images into tokens, followed by training a Transformer \cite{transformer} model using the Next Token Prediction (NTP) task to generate images. 
Recent research has explored two major directions: One is Visual autoregressive (VAR) models \cite{var,infinity}. VAR models adopt a next-scale prediction framework that significantly improves image quality. The other is multi-modal unified AR models \cite{team2024chameleon,lumina_mgpt,lumina2}. Chameleon \cite{team2024chameleon} employs an early-fusion strategy to mix text and image tokens into a single sequence, enabling unified text-to-image, captioning, and inpainting generation. Lumina‑mGPT \cite{lumina_mgpt} further incorporates progressive FP-SFT, substantially enhancing high-resolution image synthesis. Lumina‑mGPT 2.0 \cite{lumina2} approaches state-of-the-art diffusion models within a pure AR framework while maintaining strong multi-task generalization.
These unified multi-modal AR models demonstrate strong potential as a foundation for versatile and scalable vision-language generation systems.
 \begin{figure*}[!t]
    \centering
    \includegraphics[width=\linewidth]{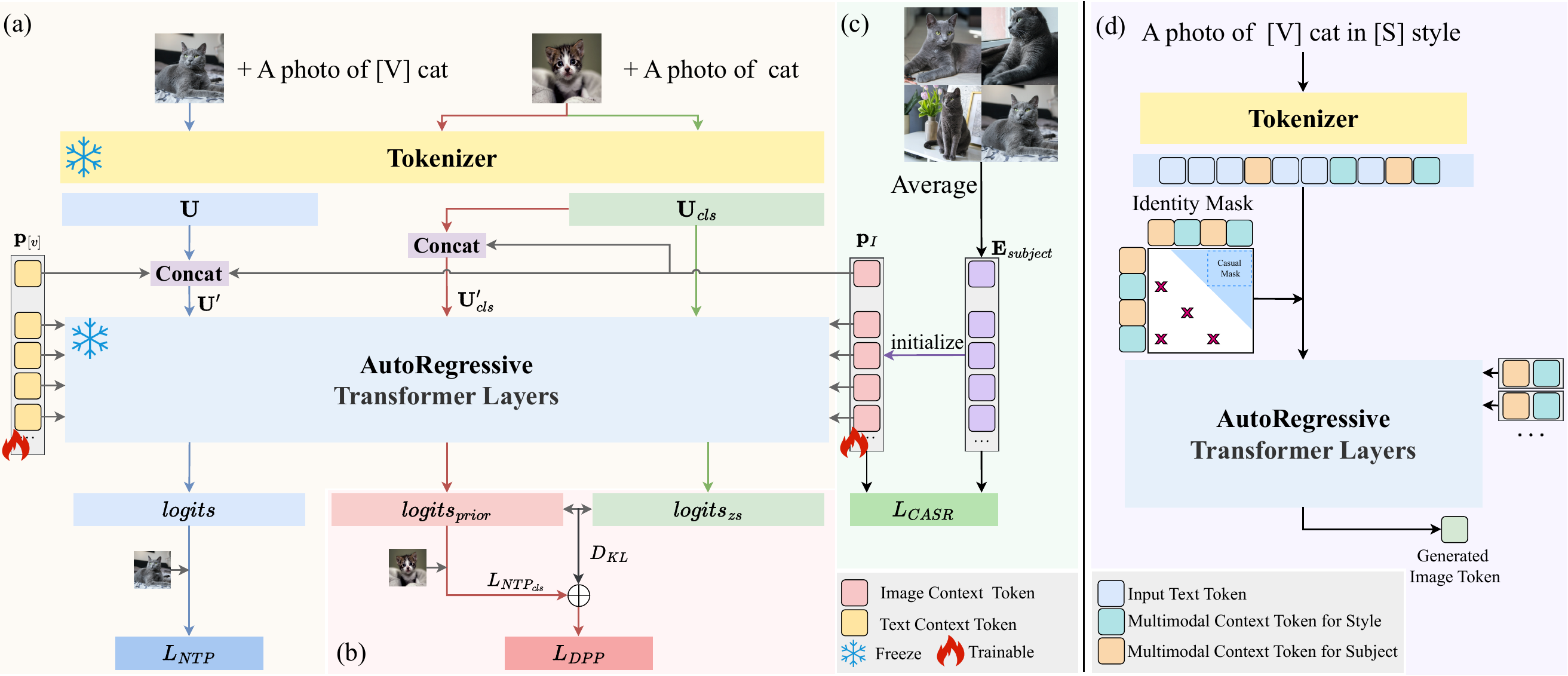}
    \captionof{figure}{ \textbf{Overview of the proposed DCoAR framework for subject-driven personalization in multi-modal autoregressive models.} (a) Layerwise Multimodal Context Learning, where learnable context tokens are injected into multiple Transformer layers for concept representation. (b) Dual Prior Preservation (DPP) regularizes the customized distribution against the pre-trained model to mitigate overfitting and language drift. (c) Context-Aware Self-Regularization (CASR) initializes and constrains context tokens towards the subject embedding space to enhance fidelity and re-contextualization. (d) Training-free subject–style composition by directly combining subject and style tokens to enable flexible customized generation.
    }
    \label{fig:framework}
\end{figure*}
\subsection{Personalized Image Generation} 
Personalized image generation aims to capture a specific concept from a few reference images and render it in novel contexts. While this task has been extensively studied in diffusion models \cite{ruiz2023dreambooth,chen2023disenbooth,han2023svdiff,customdiff,pplus_TI,textualinversion}, its application to unified autoregressive (AR) models presents unique challenges and has led to the emergence of two primary paradigms.
The first is the \textbf{adaptation-based paradigm}, which employs parameter-efficient fine-tuning (PEFT) like LoRA \cite{hu2022lora} to adapt a model's weights for a new concept \cite{personalAR, wu2025proxytuning,ARBooth,freelora}. While capable of high fidelity, this approach is hampered by overfitting and scalability bottlenecks for large-scale deployment.
The second is the \textbf{concept-injection paradigm}, which inserts concepts into a frozen model via learnable tokens to ensure scalability. Pioneering works in this area, such as Yo'Chameleon \cite{yoChameleon} and UniCTokens \cite{UniCTokens}, have explored different token designs. However, these methods are fundamentally constrained by a shallow injection strategy, where confining the concept to the input layer limits the propagation of visual details, resulting in a fidelity gap compared to adaptation-based approaches.
In contrast, DCoAR advances this paradigm by introducing a deep injection strategy. Through Layer-wise Multimodal Context Learning (LMCL), concept tokens are injected across multiple transformer layers, ensuring robust feature propagation. This approach effectively bridges the fidelity gap with adaptation-based methods while preserving the scalability inherent to the concept-injection paradigm.

\section{Method}


In this work, we propose DCoAR, a simple yet effective framework for subject-driven customization in unified autoregressive (AR) models.
The overall DCoAR framework is shown in Fig.~\ref{fig:framework}, consisting of:
\begin{itemize}
    \item \textbf{Layer-wise Multi-modal Contextual Learning}: injecting learnable multimodal tokens into multiple Transformer layers.
    \item \textbf{Dual Prior Preservation}: regularizing training with the frozen pre-trained model.
    \item \textbf{Context-Aware Self-Regularization}: initializing and constraining context tokens with subject embeddings to prevent overfitting.
\end{itemize}

\subsection{Preliminary}
In unified autoregressive (AR) text-to-image generation, both text and images are represented as sequences of discrete tokens drawn from a shared vocabulary $\mathcal{V}$. A text prompt is encoded as:
\begin{equation}
\mathbf{y} = \{\mathbf{y}_1, \mathbf{y}_2, \dots, \mathbf{y}_L\}, \quad \mathbf{y}_l \in \mathcal{V},
\end{equation}

and the target image is represented as
\begin{equation}
    \mathbf{x} = \{\mathbf{x}_1, \mathbf{x}_2, \dots, \mathbf{x}_T\}, \quad \mathbf{x}_t \in \mathcal{V},
\end{equation}
where $L$ and $T$ denote the lengths of the text and image token sequences, respectively.

AR model generates the image autoregressively by iteratively sampling the next token conditioned on the text and all previously generated tokens:
\begin{equation}
    \mathbf{x}_t \sim p_\theta(\mathbf{x}_t \mid \mathbf{x}_{<t}, \mathbf{y}), \quad t = 1, \dots, T,
\end{equation}
where $\mathbf{x}_{<t} = \{\mathbf{x}_1, \dots, \mathbf{x}_{t-1}\}$ and $\theta$ are the model parameters. 
Training follows the next-token prediction paradigm, minimizing the negative log-likelihood of the ground-truth token sequence conditioned on the text prompt:
\begin{equation}
    \mathcal{L}_{\text{NTP}} = - \sum_{t=1}^{T} \log p_\theta(\mathbf{x}_t \mid \mathbf{x}_{<t}, \mathbf{y}),
    \label{eq:ntploss}
\end{equation}
where $\mathbf{x}_t$ denotes the token at time step $t$, $\mathbf{x}_{<t}$ represents the preceding tokens, and 
$y$ is the input text prompt. This objective is implemented using the standard cross-entropy loss over the shared vocabulary $\mathcal{V}$.



\subsection{Layerwise Multimodal Context Learning Strategy}

We introduce a Layerwise Multimodal Contextual Learning (LMCL) strategy to capture the target subject concept while keeping the pre-trained model entirely frozen, requiring only a small number of learnable parameters for effective customization. 
Specifically, for all images from the same subject, we define a shared set of multimodal learnable tokens as the \textbf{representation of the subject concept}:
\begin{equation}
\mathbf{P} = \{ \mathbf{p}_{{[v]}}, \mathbf{p}_{{I}} \},
\end{equation}
where $\mathbf{p}_{{[v]}} \in \mathbb{R}^{1 \times N \times D}$ denotes the learnable tokens for the text modality with the number of insertion layers $N$ and the token embedding dimension $D$. Similarly, $\mathbf{p}_{{I}}$ represents the learnable tokens for the image modality with the same shape $1 \times N \times D$.

For each subject image, we construct a textual template using a shared prompt $\text{a photo of [V] [Class]}$, where [V] is a placeholder representing the subject identity and [Class] is the class name.
Given the original token sequence at the $i_{th}$ Transformer layer $\mathbf{U}_i$ (omitting special start/end tokens for simplicity):
\begin{equation}
\mathbf{U}_i = \{ \mathbf{y}_1, \mathbf{y}_2, \dots, \mathbf{y}_L,\mathbf{x}_1, \mathbf{x}_2, \dots, \mathbf{x}_T \}.
\end{equation}
As shown in \cref{fig:framework}(a), we insert the learnable tokens as:
\begin{equation}
\resizebox{0.9\linewidth}{!}{$\displaystyle
\mathbf{U}_i^\prime = \{ \mathbf{y}_1, \mathbf{y}_2, \dots, \mathbf{p}_{[v]}^{(i)}, \dots, \mathbf{y}_L, \mathbf{p}_{I}^{(i)}, \mathbf{x}_1, \mathbf{x}_2, \dots, \mathbf{x}_T \},
$}
\end{equation}
where $\mathbf{p}_{{[v]}}^{(i)}$ replaces the placeholder ${\text{[V]}}$ in the text sequence, and $\mathbf{p}_{{I}}^{(i)}$ is inserted before the first image token. 
Finally, we optimize the representation of these multimodal tokens using Eq.~(\ref{eq:ntploss}) to \textbf{progressively inject} the concept of the specific subject into them.

By learning only $\{\mathbf{p}_{[v]}, \mathbf{p}_{I}\}$ while keeping the backbone frozen, LMCL achieves efficient subject-driven customization with minimal parameter overhead.



\subsection{Dual Prior Preservation}
To mitigate language drift and encourage generation diversity, we propose Dual Prior Preservation (DPP) to make the prior preservation introduced in DreamBooth \cite{ruiz2023dreambooth} compatible with training of the multimodal AR model.
We first generate a small set (6–8) of class images using the pre-trained model with the prompt ``a photo of a [Class]''. These generated images, along with their corresponding text prompts, are tokenized into an input sequence $\mathbf{U}_{{cls}}$. We then perform a dual forward pass on $\mathbf{U}_{{cls}}$:
\begin{itemize}
    \item Forwarding the original $\mathbf{U}_{{cls}}$ through the pretrained model to obtain a zero-shot output distribution ${\text{logits}}_{{zs}}$.
    \item Inserting layerwise image context tokens into $\mathbf{U}_{{cls}}$ yields $\mathbf{U}_{{cls}}'$ and forward through the current model to obtain $\text{logits}_{{prior}}$. 
\end{itemize}

The DPP loss is computed as:
\begin{equation}
    \begin{aligned}
      \mathcal{L}_{{DPP}} &= \lambda_1 \cdot \mathcal{L}_{{NTP_{cls}}}({\text{logits}}_{{prior}}, \text{labels}_{cls}) \\
      & + \lambda_2 \cdot D_{{KL}}({\rm{logits}}_{{zs}} \parallel {\rm{logits}}_{{prior}}), \\
    \end{aligned}
    \label{eq:dpp_loss}
\end{equation}
where $\mathcal{L}_{{NTP_{cls}}}$ denotes the Next Token Prediction (NTP) loss on the class images in Eq.~(\ref{eq:ntploss}),
$\text{labels}_{cls}$ are the ground-truth visual tokens of the class images, $D_{{KL}}$ is the Kullback–Leibler (KL) divergence measuring the distance between the two output distributions, which is formulated as:
\begin{equation}
    \begin{aligned}
D_{{KL}}({\rm{logits}}_{{zs}} \parallel {\rm{logits}}_{{prior}}) 
&= \\ \sum_{v \in \mathcal{V}} {\rm{logits}}_{{zs}}(v) \log& \frac{ {\rm{logits}}_{{zs}}(v)}{{\rm{logits}}_{{prior}}(v)}
    \end{aligned},
\end{equation}
and $\lambda_1, \lambda_2$ are coefficients balancing the NTP and KL regularization terms.

The dual-path design extends DreamBooth’s prior preservation from continuous diffusion latents to discrete token modeling, making it compatible with multi-modal AR frameworks.

\subsection{Context-Aware Self-Regularization}
To address overfitting and improve re-contextualization in subject-driven customization, we propose the Context-Aware Self-Regularization (CASR).
As shown in \cref{fig:framework}(c), given the reference images of the specific subject, we feed all samples into the frozen pre-trained model and compute the average embedding of all image tokens in each layer $i$, denoted as $\mathbf{E}_{{subject}}^{(i)}$. We initialize the learnable image context token $\mathbf{p}_{I}^{(i)}$ with $\mathbf{E}_{{subject}}^{(i)}$, enabling training to start from a representation closer to the subject’s distribution and accelerating convergence.
The CASR loss is defined as:
\begin{equation}
    \mathcal{L}_{{CASR}} = \frac{1}{N}\sum_{i=1}^{N} \big\| \mathbf{p}_{I}^{(i)} - \mathbf{E}_{{subject}}^{(i)} \big\|^2 ,
\label{eq:casrloss}
\end{equation}
where $\|\cdot\|^2$ denotes the squared $\ell_2$-norm measuring the Euclidean distance and $N$ is the total number of insertion layers.

CASR regularizes the context tokens by constraining them towards the subject’s embedding space, effectively preventing overfitting and significantly improving re-contextualization performance. 




\subsection{Overall Objective}
The final training objective combines the autoregressive next-token prediction with the proposed regularization terms:
\begin{equation}
    \mathcal{L}_{obj} = \mathcal{L}_{{NTP}} + \alpha \mathcal{L}_{{DPP}} + \beta \mathcal{L}_{{CASR}},
\end{equation}
where $\mathcal{L}_{{NTP}}$ denotes the next-token prediction loss on the subject images, $\alpha$ and $\beta$ are two factors that balance the contribution of the DPP and CASR losses, respectively.

\begin{figure*}[!t]
    \centering
    \includegraphics[width=\linewidth]{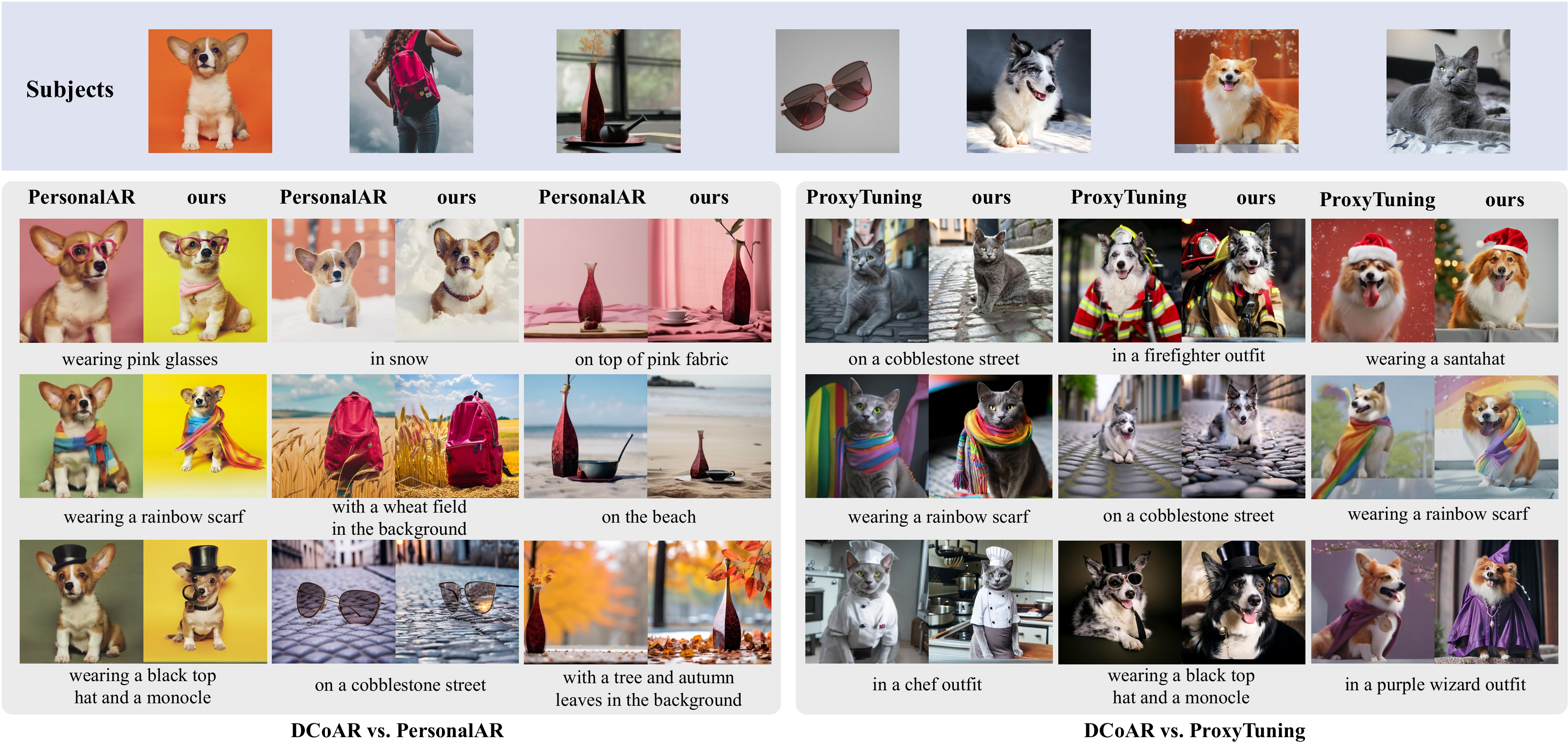}
    \vspace{-0.6cm}
    \captionof{figure}{Qualitative comparison of subject-driven personalization on the DreamBench benchmark. As concept-injection methods generally exhibit lower visual fidelity, we focus the qualitative comparison on the more competitive adaptation-based methods. \textbf{Additional comparisons are available in the supplementary material.}}
    \label{fig:visualization_subject}
\end{figure*}

\begin{table*}[!t]
    \centering
    \begin{tabular}{@{}l l c c c c@{}}
        \toprule
        \textbf{Method Tyeps} & \textbf{Method} & \textbf{DINO} & \textbf{CLIP-I} & \textbf{CLIP-T} & \textbf{Trainable Params} \\
        \midrule
         & Real Imges & 0.774 & 0.885 & N/A & N/A \\
        \midrule
        \multirow{4}{*}{Diffusion Models}  
        & TI\cite{textualinversion} & 0.569 & 0.780 & 0.255 & N/A \\
        & DreamBooth(SD)\cite{ruiz2023dreambooth} & 0.668 & 0.803 & 0.305 & N/A \\
        & DreamBooth (Imagen)\cite{ruiz2023dreambooth} & 0.696 & 0.812 & 0.306 & N/A \\
        & DreamBooth (Flux)\cite{wu2025proxytuning} & 0.732 & 0.802 & 0.314 & N/A \\
        \midrule
        \multirow{8}{*}{Unified AR Models} 
        & DreamBooth(Lumina-mGPT)\cite{wu2025proxytuning} & 0.509 & 0.675 & 0.295 & N/A \\
        & DreamBooth(Lumina-mGPT)\cite{wu2025proxytuning} +LoRA & 0.534 & 0.697 & 0.265 & N/A \\
        & Yo'Chameleon\cite{yoChameleon} & 0.542 & 0.795 & 0.225 & 0.13M \\
        & UniCTokens\cite{UniCTokens} & 0.599 & 0.782 & 0.304 & 0.13M \\
        & PersonalAR\cite{personalAR} & 0.671 & 0.805 & 0.302 & 1610.6M \\
        & PersonalAR\cite{personalAR} +LoRA & 0.657 & 0.778 & 0.312 & 50.4M \\
        & Proxy-Tuning\cite{wu2025proxytuning} & \textbf{0.752} & 0.809 & 0.312 & 142.6M \\
        & \textbf{DCoAR(Ours)} & 0.723 & \textbf{0.815} &\textbf{0.318} & \textbf{0.073M} \\
        \bottomrule
    \end{tabular}
    \caption{Performance comparison with state-of-the-art methods on the DreamBench dataset. The results for competing methods are directly cited from previous publications and \textbf{details on parameter count calculation can be found in the supplementary material.}}
    \label{tab:performance_comparison}
\end{table*}

\subsection{Customize Any Subject in Any Style}
Since DCoAR leaves the pre-trained AR model parameters untouched, it fully leverages the strong multimodal understanding and generative capacity of autoregressive models, which naturally support zero-shot subject–style composition. 
This design enables training-free customization of arbitrary subjects in arbitrary styles simply by concatenating their respective context tokens, allowing for flexible, efficient, and scalable rendering without fine-tuning.
To further ensure clean and controllable generation, we introduce an Identity Mask during inference, as shown in \cref{fig:framework}(d), which explicitly restricts attention flow between subject and style tokens.
This mechanism prevents semantic interference and ensures that each token type contributes distinctly and independently to the final generation.




\section{Experiments}
\subsection{Experimental Details}
\begin{figure}[!t]
    \centering
    \includegraphics[width=1\linewidth]{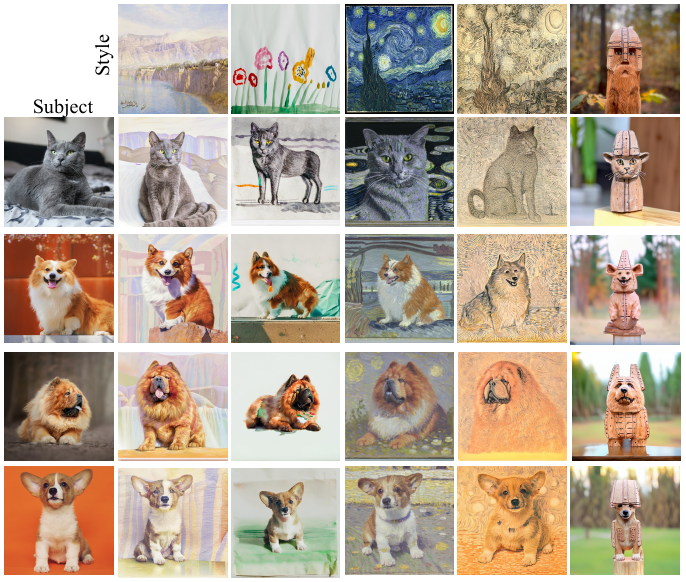}
    \caption{Training-free subject-style generation with our DCoAR on DreamBench \cite{ruiz2023dreambooth} and StyleDrop \cite{sohn2023styledrop} datasets.}
    \label{fig:style_recon}
\end{figure}
\begin{figure}[!t]
    \centering
    \includegraphics[width=1\linewidth]{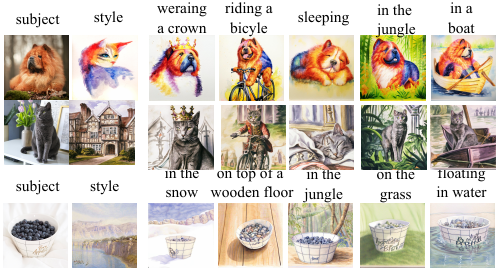}
    \caption{Re-contextualization with our DCoAR on DreamBench \cite{ruiz2023dreambooth} and StyleDrop \cite{sohn2023styledrop} datasets.}
    \label{fig:style_recon}
\end{figure}



\textbf{Benchmark.} We evaluate subject-driven customization using the DreamBench dataset \cite{ruiz2023dreambooth}, which contains 30 subjects (21 objects, 9 animals), each with 3–5 reference images and 25 test prompts.
Objects are tested with 20 recontextualization and 5 property modification prompts, while animals use 10 recontextualization, 10 accessorization, and 5 property modification prompts.
For stylization, we adopt the StyleDrop dataset \cite{sohn2023styledrop} and follow ZipLoRA \cite{shah2024ziplora}, training with a single reference image per style.
Additionally, we evaluate subject-style hybrid generation by combining DreamBench subjects with StyleDrop styles.

\noindent
\textbf{Implementation Details.}
All experiments are run on a single H800 GPU using Lumina-mGPT-7B FP-SFT \cite{lumina_mgpt} as the backbone. For subject-driven personalization, we insert one text and one image token per layer into the first 9 layers. For stylization with a single reference image, the same token configuration is applied to the first 3 layers.
The model is trained with a learning rate of 1e-2 and batch size 1, keeping all other Lumina-mGPT hyperparameters unchanged. We set $\alpha$ = 1e-2, $\beta$ = 5e-4, and both $\lambda_1$ and $\lambda_2$ to 0.5. Each subject and style is trained for 1000 and 600 steps, respectively.

\begin{table}[t]
    \centering
\resizebox{0.99\linewidth}{!}{
    \begin{tabular}{l c c c}
        \toprule
        & \textbf{DCoAR (Ours)} & B-LoRA & ZipLoRA \\
        \midrule
        Subject-alignment & 0.604 & 0.579 & \textbf{0.655}\\
        Style-alignment & \textbf{0.605} & 0.505 & 0.597\\
        Text-alignment & \textbf{0.308} & 0.258 & 0.272\\
        \bottomrule
    \end{tabular}
    }
    \caption{Comparison results of our DCoAR and two competitors on the style personalization task. }
    \label{tab:style-comparison}
\end{table}

\noindent
\textbf{Evaluation Metrics.}
For subject-driven customization tasks, we follow established protocols and employ CLIP-I \cite{clip} and DINO \cite{dino} for subject fidelity evaluation, along with CLIP-T \cite{clip} for prompt fidelity assessment. Specifically, CLIP-I computes the average cosine similarity between CLIP embeddings of real and generated images. DINO measures the average cosine similarity of ViT-S/16 DINO embeddings between real and generated images. CLIP-T evaluates the cosine similarity between CLIP text embeddings of prompts and CLIP image embeddings of generated outputs. For stylization tasks, we adopt metrics from ZipLoRA \cite{shah2024ziplora}, which use CLIP features for both style and text alignment, and DINO features for subject alignment.

\noindent
\textbf{Competitors.} For subject-driven personalization on the DreamBench dataset, we compare our approach with several baselines. These include diffusion-based methods like Textual Inversion (TI) \cite{textualinversion} and DreamBooth \cite{ruiz2023dreambooth}. We also include a range of Unified Multimodal Models (MLLMs) that share an autoregressive architecture similar to ours. This category features DreamBooth adapted for MLLMs, Yo'Chameleon\cite{yoChameleon}, UniCTokens\cite{UniCTokens}, PersonalAR \cite{personalAR}, and Proxy-Tuning \cite{wu2025proxytuning}. For style personalization, we compare with recent diffusion-based methods: ZipLoRA \cite{shah2024ziplora} and B-LoRA \cite{b-lora}.

\subsection{Main Results}
\textbf{Quantitative Results of Subject-Driven Personalization.} Table \ref{tab:performance_comparison} shows the comparison results of our DCoAR and the competitors on the DreamBench dataset. 
From the results, we have the following observations.
First, DCoAR achieves superior performance across multiple dimensions. In the CLIP-I metric, DCoAR achieves a leading score of 0.8151, surpassing the second-best DreamBooth (Imagen) by 0.4\%, demonstrating its exceptional image fidelity. For CLIP-T, DCoAR sets a new benchmark at 0.3184, outperforming the second-best method, Proxy-Tuning by 0.29\%, highlighting its superior cross-modal alignment capability. While DCoAR ranks second on the DINO metric, it is substantially more efficient and user-friendly than the top-performing method, Proxy-Tuning, which requires training an additional diffusion model and generating hundreds to a thousand images per subject for data augmentation.

\noindent
\textbf{Quantitative Results of Style Personalization.} Table~\ref{tab:style-comparison} presents a comparison between our DCoAR model and two baselines: B-LoRA~\cite{b-lora} and ZipLoRA (SDXL)~\cite{shah2024ziplora}. The results demonstrate that DCoAR outperforms B-LoRA across all three evaluation metrics, achieving the highest text-alignment score among all competitors. While DCoAR trails ZipLoRA slightly in subject alignment (0.604 vs. 0.655), it achieves comparable style preservation (0.605 vs. 0.597), superior text alignment (0.308 vs. 0.272), and does so without requiring any additional training.

\noindent
\textbf{Qualitative Results of Subject-Driven Personalization.}  
\cref{fig:visualization_subject} illustrates some generated samples of our DCoAR compared with two most related competitors, i.e., PersonalAR \cite{personalAR} and Proxy-Tuning \cite{wu2025proxytuning}. From the results, we observe that images generated by DCoAR exhibit superior prompt fidelity. For instance, given the instruction ``wearing a black top hat and a monocle", neither PersonAR nor Proxy-Tuning successfully generates the ``monocle", whereas our DCoAR does. Moreover, our generated images better adhere to physical plausibility. In the second row of PersonAR, the generated vase and backpack are both floating, which is unrealistic, while such artifacts are absent in our results. 
More qualitative results are provided in the supplementary materials.
\begin{figure}[!t]
    \centering
    \includegraphics[width=1\linewidth]{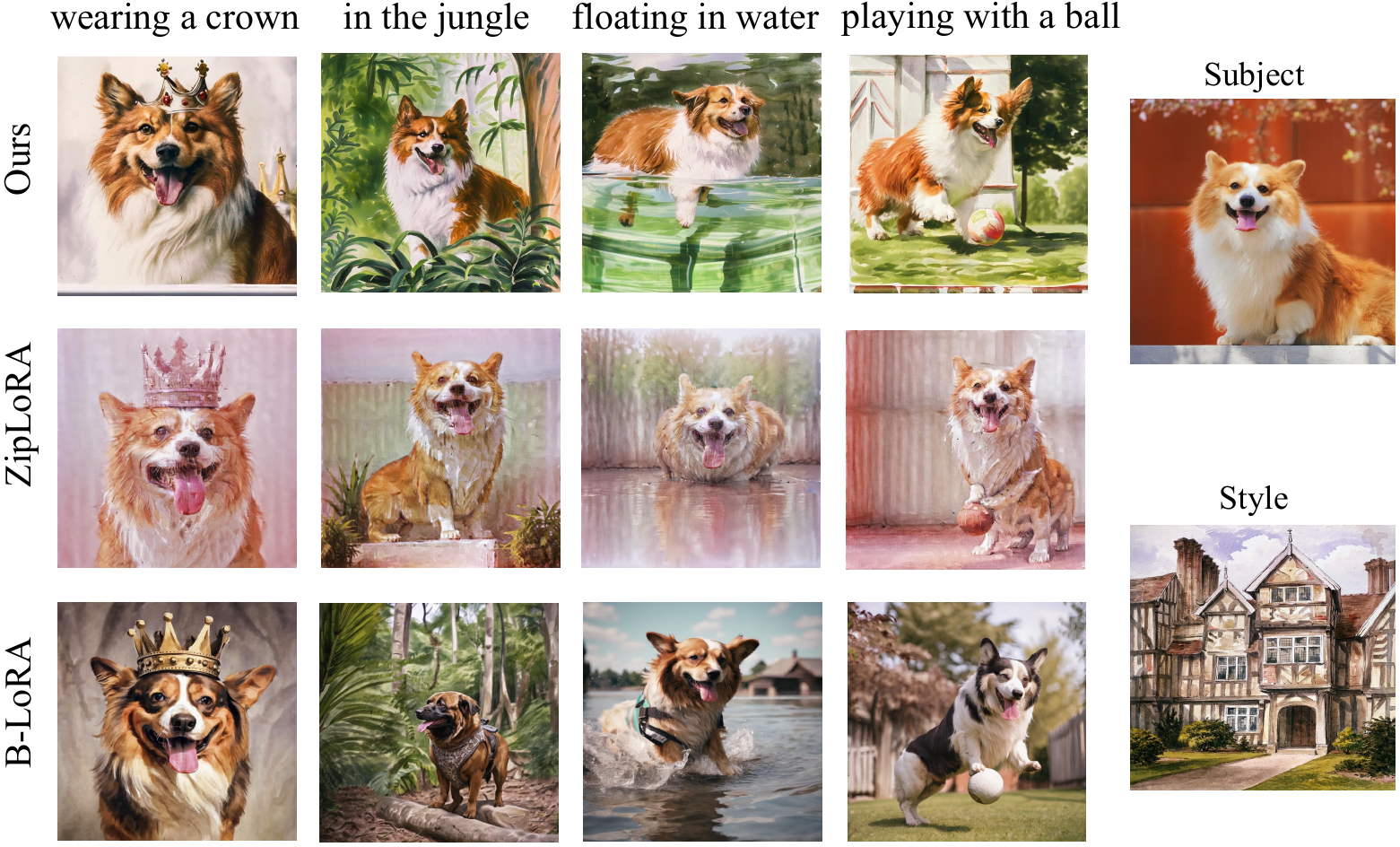}
    \caption{Qualitative comparison of our DCoAR and LoRA-based method for style personalization tasks.}
    \label{fig:style_compare}
\end{figure}

\noindent
\textbf{Qualitative Results of Style Personalization.}
As shown in \cref{fig:style_compare}, DCoAR demonstrates superior performance compared to B-LoRA.
While ZipLoRA preserves more subject details, such as color, DCoAR exhibits stronger adherence to instructions when generating in novel scenes (e.g., ``in the jungle'' or ``floating in water'').
Additional qualitative results on style customization are presented in \cref{fig:style_recon}, showing that DCoAR effectively preserves the given style when generalizing the reference subject to novel scenes.

\subsection{Ablation Studies} 
\begin{figure}
    \centering
    \includegraphics[width=0.99\linewidth]{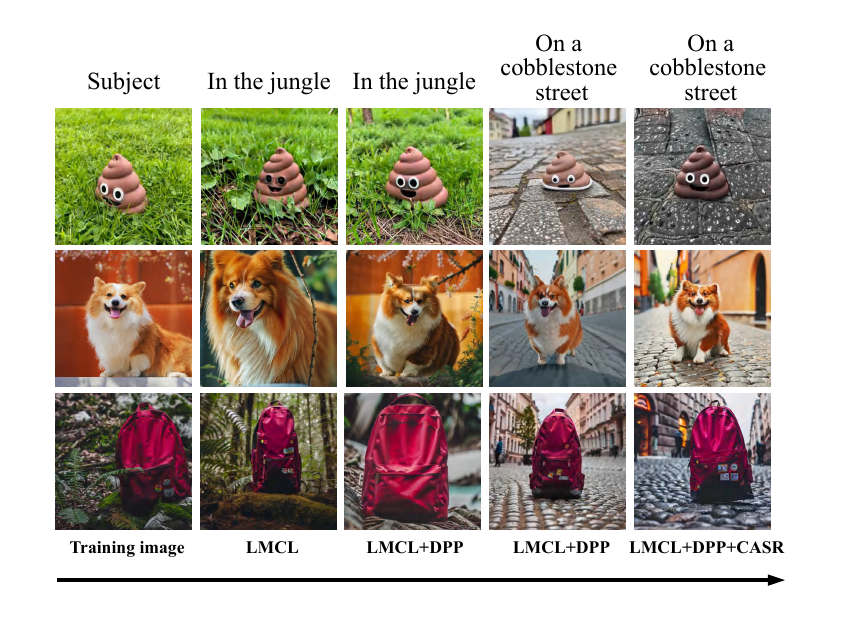}
    \caption{Visualization of the effects of different losses on the generated samples. }
    \label{fig:ablation_result}
\end{figure}

\textbf{Impacts of Different Losses.} We perform an ablation study by incrementally applying the LMCL, DPP, and CASR losses and evaluating performance on DINO, CLIP-I, and CLIP-T metrics, as shown in Tab.~\ref{tab:loss-ablation}. Only applying LMCL yields baseline scores of 0.6610 (DINO), 0.7647 (CLIP-I), and 0.3096 (CLIP-T).
Introducing DPP leads to consistent improvements across all metrics, increasing DINO to 0.7142 and CLIP-I to 0.8019. Although CLIP-T slightly decreases to 0.2977, the overall alignment benefits are clear. Alternatively, adding the CASR loss to the LMCL baseline improves all metrics, with scores of 0.7194 (DINO), 0.7905 (CLIP-I), and 0.3192 (CLIP-T).
When all three losses (LMCL, DPP, and CASR) are applied together, we observe the best overall performance, achieving the highest scores for DINO (0.7226) and CLIP-I (0.8151), and a strong result for CLIP-T (0.3184). This confirms that each loss component contributes positively to the overall alignment quality and subject fidelity.
As shown in \cref{fig:ablation_result}, we analyze the results using only Layer-wise Multi-modal Contextual Learning (LMCL), and then with the addition of the Dual Prior Preservation (DPP) loss. With the DPP constraint, the model better preserves the subject’s identity. However, when recontextualizing the subject in the scene "on a cobblestone street", it still struggles to accurately render the background. In contrast, incorporating Context-Aware Self-Regularization (CASR) enables the model to successfully generalize to the new scene.
\begin{table}[t]
    \centering
    \footnotesize
    \resizebox{0.95\linewidth}{!}{
    \begin{tabular}{cccccc}
        \toprule
        LMCL & DPP & CASR & DINO & CLIP-I & CLIP-T \\
        \midrule
        \checkmark & & & 0.6610 & 0.7647 & 0.3096 \\
        \checkmark & \checkmark & & 0.7142 & 0.8019 & 0.2968 \\
        \checkmark & & \checkmark &  0.7194 & 0.7905 & \textbf{0.3192} \\
        \checkmark & \checkmark & \checkmark & \textbf{0.7226} & \textbf{0.8151} & 0.3184 \\
        \bottomrule
    \end{tabular}}
    \caption{Impacts of different losses on the subject-driven personalization task.}
    \label{tab:loss-ablation}
\end{table}
\begin{figure}[t]
    \centering
    \includegraphics[width=0.96\linewidth]{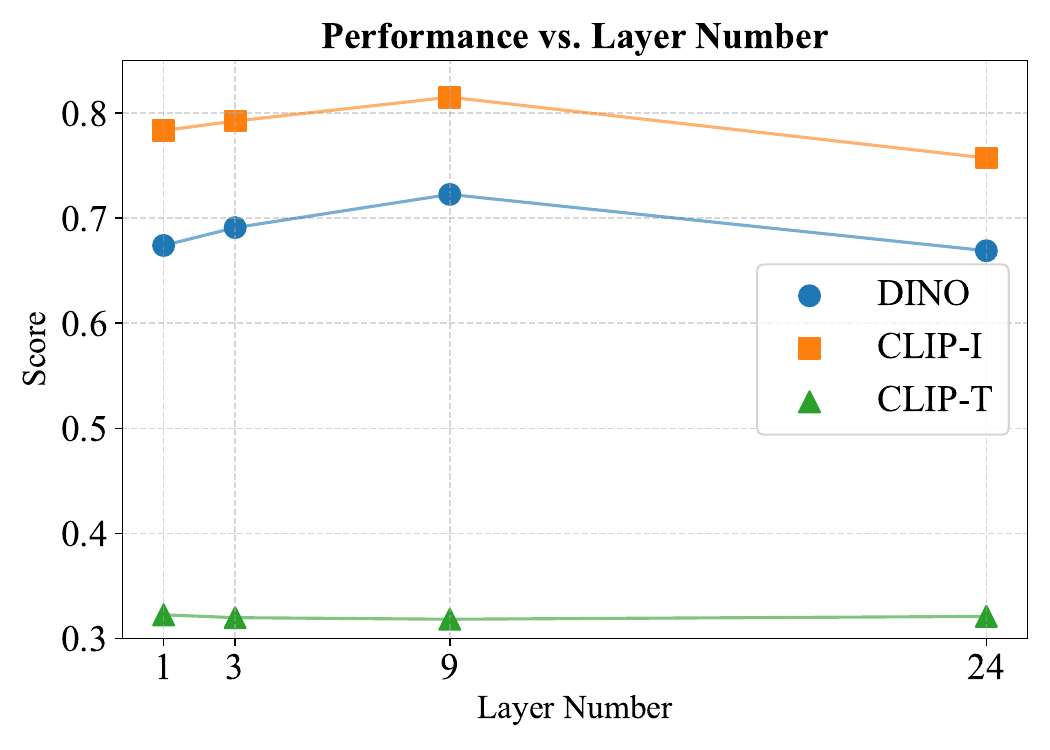}
    \caption{Impacts of insertion depth of multimodal context tokens on performance. }
    \label{fig:ablation_layernum}
\end{figure}

\noindent
\textbf{Impacts of Layer Numbers.}
To evaluate the impacts of depth of multimodal context token insertion, we insert tokens at layers 1, 3, 9, and 24. As illustrated in \cref{fig:ablation_layernum}. CLIP-T remains relatively stable across depths, while CLIP-I and DINO follow a rise-then-fall trend. Notably, inserting tokens at deeper layers (e.g., layer 24) leads to a decline in subject fidelity. We attribute this to overfitting caused by limited reference data, which becomes more pronounced as token influence shifts to higher-level semantic layers.

\begin{figure}
    \centering
    \includegraphics[width=0.99\linewidth]{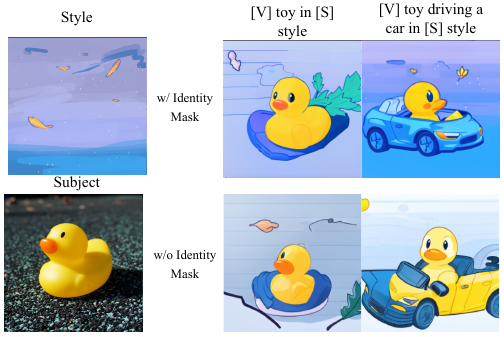}
    \caption{Ablation results of Identity Mask. }
    \label{fig:ablation_mask}
\end{figure}

\noindent
\textbf{Impacts of Identity Mask.}
\Cref{fig:ablation_mask} demonstrates the critical role of our Identity Mask. Without it, the model suffers from severe concept contamination, causing the subject's color to erroneously bleed into other scene elements. The mask prevents this by restricting attention flow between subject and style tokens, enforcing a clean disentanglement that ensures semantically coherent compositional generation.
We provide more detailed analysis and results in the supplementary material.

\section{Conclusion} \label{sec:conclusion}
In this paper, we introduced DCoAR, a framework that advances the concept-injection paradigm for AR model personalization by replacing previous shallow strategies with deep concept injection. Our method, Layer-wise Multimodal Context Learning, ensures robust feature propagation while keeping the backbone model completely frozen. Experiments demonstrate that DCoAR successfully bridges the fidelity gap, achieving the high performance while retaining superior parameter-efficiency. Moreover, its frozen backbone enables a training-free stylization strategy competitive with modern diffusion methods.

\clearpage
\onecolumn
\setcounter{page}{1}
\begin{center}
   \Large \textbf{DCoAR: Deep Concept Injection into Unified Autoregressive Models for Personalized Text-to-Image Generation} \\[0.5em]
   \large \textbf{Supplementary Material} \\[0.5em]
\end{center}

\section{Training and Evaluation Details}
\textbf{Training Details.} The training details of DCoAR on different datasets are shown in \Cref{Table:training_details}. For the style personalization task, only the provided single reference image is utilized. Consequently, the hyperparameters for Dual Prior Preservation (DPP) are not applied.

\noindent
\textbf{Evaluation Details.} For the subject-style generation task, we selected 8 subjects from DreamBench \cite{ruiz2023dreambooth}, including 4 animals and 4 objects, as well as 6 styles from StyleDrop  \cite{sohn2023styledrop}. These yielded a total of 48 unique subject-style combinations, on which we evaluated our DCoAR, ZipLoRA \cite{shah2024ziplora}, and B-LoRA \cite{b-lora}. For each combination, we assigned 10 distinct recontextualization prompts for both objects and living subjects, as illustrated in \Cref{Table:prompt_details}.

\begin{table}[h]
    \centering
    \begin{tabular}{lccc}
    \toprule
     & DreamBench & StyleDrop \\
    \midrule
    lr & 1e-2 & 1e-2 \\
    batch size & 1 & 1 \\
    resolution & 768 & 768 \\
    $\lambda_1$ & 0.5 & N/A  \\
    $\lambda_2$ & 0.5 & N/A \\
    $\alpha$ & 1e-2 & N/A  \\
    $\beta$ & 5e-4 & 5e-4  \\
    Training Steps & 1000 & 600 \\
    Context token layers & 9 & 3 \\
    \bottomrule
    \end{tabular}
    \caption{Training settings of DCoAR on DreamBench\cite{ruiz2023dreambooth} and StyleDrop \cite{sohn2023styledrop} datasets.}
    \label{Table:training_details}
\end{table}

\begin{table}[h]
    \centering
    \begin{tabular}{ll @{\hspace{1cm}} ll} 
    \toprule
    \textbf{Subject} & \textbf{Context} & \textbf{Subject} & \textbf{Context} \\
    \midrule
    \multirow{10}{*}{Objects} 
      & in the snow                   & \multirow{10}{*}{Living  Subjects} & wearing a hat \\
      & on the beach                  & & with a crown \\
      & with a city in the background & & riding a bicycle \\
      & on top of a dirt road         & & sleeping \\
      & on top of green grass         & & in a boat \\
      & in the jungle                 & & in the jungle \\
      & on the mountain               & & on the mountain \\
      & floating in water             & & floating in water \\
      & on a picnic table             & & on a picnic table \\
      & on top of a wooden floor      & & playing with a ball \\
    \bottomrule
    \end{tabular}
    \caption{Prompt templates categorized by subject type.}
    \label{Table:prompt_details}
\end{table}

\section{Details of Parameter Calculation}
For PersonalAR \cite{personalAR}, we directly reference the parameter count reported in its original paper. As for Proxy-Tuning \cite{wu2025proxytuning}, since the paper does not provide explicit parameter statistics and the code is not publicly available, we provide a coarse estimate based on the hyperparameter configurations described in the paper.
Proxy-Tuning is a two-stage approach where a diffusion model is first trained, followed by an autoregressive (AR) model that is trained on data generated by the diffusion model. Both models are fine-tuned using Low-Rank Adaptation (LoRA) \cite{hu2022lora}. In our estimation, the diffusion model is based on Stable Diffusion 3.5 (SD3.5 Large) \cite{sd3}, while the AR model is Lumina-mGPT 7B \cite{lumina_mgpt}.
We estimate the total number of trainable LoRA parameters based on the typical LoRA application settings.

\subsection{LoRA Parameter Calculation}

For a weight matrix \( W \in \mathbb{R}^{d \times d} \), LoRA introduces two trainable matrices \( A \in \mathbb{R}^{d \times r} \) and \( B \in \mathbb{R}^{r \times d} \), where \( r \) is the LoRA rank. The number of trainable parameters for each projection is:

\[
\text{LoRA params} = 2dr,
\]
We assume LoRA is applied to all attention projections: \( W_q \), \( W_k \), \( W_v \), and \( W_o \).

\subsection*{Diffusion Model (SD3.5)}

SD3.5 is based on the DiT \cite{peebles2023scalablediffusionmodelstransformers} architecture, with 24 Transformer \cite{transformer} blocks. Each block includes one self-attention and one cross-attention module. Assuming a hidden size \( d = 3072 \), the LoRA (rank=64) parameters per attention module are:
\begin{equation}
    \begin{aligned}
        4 \times 2 \times d \times r &= 8dr \\
    &= 8 \times 3072 \times 64 \\
    &= 1{,}572{,}864 \approx 1.57M.
    \end{aligned}
\end{equation}
There are a total of \( 24 \times 2 = 48 \) attention modules, leading to:
\begin{equation}
    \begin{aligned}
48 \times 1.57\text{M} = 75.5\text{M}
    \end{aligned}
\end{equation}
\subsection{AR Model (Lumina-mGPT 7B)}
For the AR model, we assume a hidden size \( d = 4096 \) and 32 Transformer layers. Applying LoRA (rank = 64) to the four attention projections in each layer results in:
\begin{equation}
    \begin{aligned}
4 \times 2 \times d \times r &= \\
&=8dr \\
&= 8 \times 4096 \times 64 \\
&= 2{,}097{,}152 \approx 2.097M,
    \end{aligned}
\end{equation}
leading to,
\begin{equation}
    \begin{aligned}
32 \times 2.097\text{M} = 67.1\text{M}
    \end{aligned}
\end{equation}
\subsection{Total Parameter Count}
Summing both components, the total number of trainable LoRA parameters under Proxy-Tuning is approximately:
\begin{equation}
    \begin{aligned}
75.5\text{M} + 67.1\text{M} = 142.6\text{M}
    \end{aligned}
\end{equation}
This estimate assumes LoRA is applied to all attention projections and excludes feed-forward layers. Including FFNs would increase the total parameter count further, but standard LoRA configurations typically restrict adaptation to attention modules only.

\section{Additional Experiments}

\subsection{Ablation Study on Dual Prior Preservation (DPP)}
To validate the design of our Dual Prior Preservation (DPP) strategy, we evaluate the individual contributions of the Class-based Next Token Prediction loss ($\mathcal{L}_{NTP_{cls}}$) and the KL Divergence constraint ($D_{KL}$). As shown in Table~\ref{tab:dpp_ablation}, applying these components in isolation leads to sub-optimal performance, highlighting the necessity of their joint application.

Specifically, employing \textbf{only $\mathcal{L}_{NTP_{cls}}$} (Row 2) results in a significant drop in CLIP-T ($0.3192 \rightarrow 0.3011$). This indicates that a hard reconstruction constraint without distributional regularization causes the model to overfit the visual appearance of the reference, leading to severe language drift and impaired re-contextualization capabilities. 

Conversely, applying \textbf{only $D_{KL}$} (Row 3) maintains text alignment but causes a sharp decline in subject fidelity (DINO drops to $0.7023$, CLIP-I to $0.7794$). This suggests that the soft distribution constraint alone is too conservative, preventing the model from learning the unique, fine-grained details of the specific subject.

However, when \textbf{combined} (Row 4), the two components exhibit a strong synergistic effect. The $\mathcal{L}_{NTP_{cls}}$ term ensures high subject fidelity (boosting CLIP-I to $\mathbf{0.8151}$), while the $D_{KL}$ term effectively regularizes the distribution to prevent language drift, maintaining a high CLIP-T score of $0.3184$. This confirms that the full DPP loss is essential for balancing identity preservation and text controllability.

\begin{table}[h]
    \centering
    \begin{tabular}{cc ccc}
    \toprule
    \multicolumn{2}{c}{\textbf{Components}} & \multicolumn{3}{c}{\textbf{Metrics}} \\
    \cmidrule(lr){1-2} \cmidrule(lr){3-5}
    $\mathcal{L}_{NTP_{cls}}$ & $D_{KL}$ & \textbf{DINO} ($\uparrow$) & \textbf{CLIP-I} ($\uparrow$) & \textbf{CLIP-T} ($\uparrow$) \\
    \midrule
    & & 0.7194 & 0.7905 & 0.3192 \\
    \checkmark & & 0.7188 & 0.7893 & 0.3011 \\
    & \checkmark & 0.7023 & 0.7794 & 0.3189 \\
    \rowcolor{gray!10} \checkmark & \checkmark & \textbf{0.7226} & \textbf{0.8151} & 0.3184 \\
    \bottomrule
    \end{tabular}
    \caption{Ablation study on the individual components of the Dual Prior Preservation (DPP) loss using the DreamBench dataset.}
    \label{tab:dpp_ablation}
\end{table}

\subsection{Visualization of the effects of CASR.}
\begin{figure}
    \centering
    \includegraphics[width=0.8\linewidth]{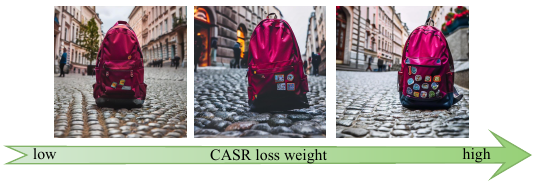}
    \caption{Visualization of CASR. }
    \label{fig:ablation_casr}
\end{figure}

\Cref{fig:ablation_casr} provides a visual illustration of the effect of our proposed CASR loss. When the loss weight is too small, the model tends to overfit: although subject fidelity remains high, generalization to novel contexts is limited. Conversely, when the weight is too large, the context tokens become under-optimized, leading to a significant degradation in subject fidelity.

\begin{figure}[t]
    \centering
    \includegraphics[width=\linewidth]{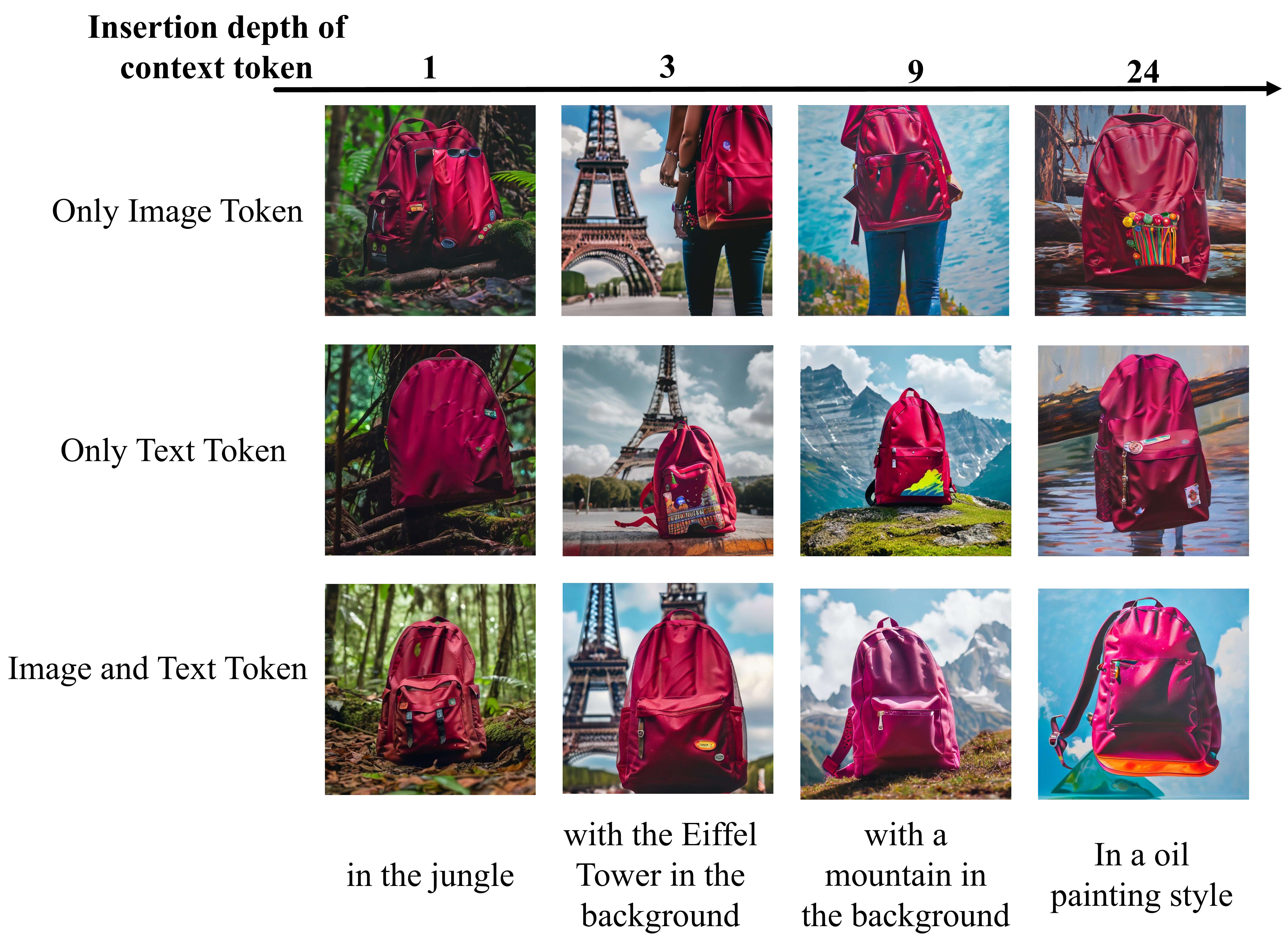}
    \caption{\textbf{Qualitative ablation of token modalities across different insertion depths.} Rows represent different token configurations, while columns correspond to the number of Transformer layers (1, 3, 9, 24) into which tokens are injected. 
    \textbf{Rows:} ``Only Text Token'' suffers from identity loss (generic backpacks), while ``Only Image Token'' exhibits high fidelity but poor editability (e.g., failing to stylize in Col 4). 
    \textbf{Columns:} Inserting tokens into all layers (Depth 24) leads to overfitting, preventing style changes, whereas the optimal depth (Depth 9) achieves the best balance between subject fidelity and textual control.}
    \label{fig:ablation_depth_modalit}
\end{figure}

\subsection{Joint Analysis of Modality and Insertion Depth}

Figure~\ref{fig:ablation_depth_modalit} provides a comprehensive visualization of how different token modalities behave across varying insertion depths (injecting tokens into the first $N$ layers, where $N \in \{1, 3, 9, 24\}$).

\noindent\textbf{Impact of Modalities (Rows).}
\begin{itemize}
    \item \textbf{Only Text Tokens ($p_{[v]}$):} As seen in the second row, relying solely on text tokens results in \textit{poor identity preservation} across all depths. Although the text alignment is high (e.g., the ``oil painting'' style in Col 4 is correctly rendered), the generated backpacks are generic and lack the fine-grained details of the reference subject.
    
    \item \textbf{Only Image Tokens ($p_I$):} The first row shows that while image tokens preserve visual details, they suffer from \textit{rigidity and poor instruction following}, especially at deeper insertion layers.
    
    \item \textbf{Image and Text Tokens (Ours):} The third row demonstrates that combining both modalities ensures robust identity preservation while remaining responsive to text prompts.
\end{itemize}

\vspace{0.5em}
\noindent\textbf{Impact of Insertion Depth (Columns).}
\begin{itemize}
    \item \textbf{Shallow Injection (Depth 1--3):} In the early columns, the subject identity is not fully consolidated. For example, at Depth 1, the backpack's texture and shape appear slightly inconsistent with the reference, indicating insufficient visual signal propagation.
    
    \item \textbf{Deep Injection (Depth 24):} The last column (Depth 24) reveals the detrimental effect of \textit{over-injection}. In the ``Only Image Token'' setting, the backpack fails to transform into the ``oil painting'' style and remains photorealistic. This indicates that injecting visual tokens into all layers creates an overly strong visual prior that overrides the textual style control, leading to \textbf{overfitting}.
    
    \item \textbf{Optimal Depth (Depth 9):} Our chosen setting (Depth 9, Column 3) achieves the optimal sweet spot. The model successfully retains the specific identity of the backpack (unlike the text-only baseline) while seamlessly integrating it into the new background (unlike the depth-24 baseline), validating our decision to inject concepts only into the early-to-mid layers.
\end{itemize}

\subsection{Additional Qualitative Results}
\begin{figure*}[htbp]
    \centering
    \includegraphics[width=0.99\linewidth]{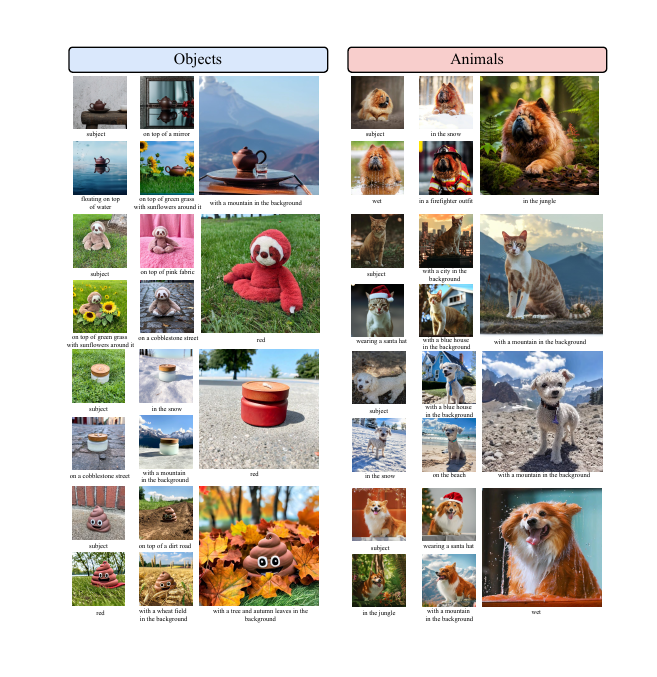}
    \caption{Qualitative results of subject-driven customization with DCoAR.}
    \label{Fig:qual_subject2}
\end{figure*}

\begin{figure*}[htbp]
    \centering
    \includegraphics[width=0.99\linewidth]{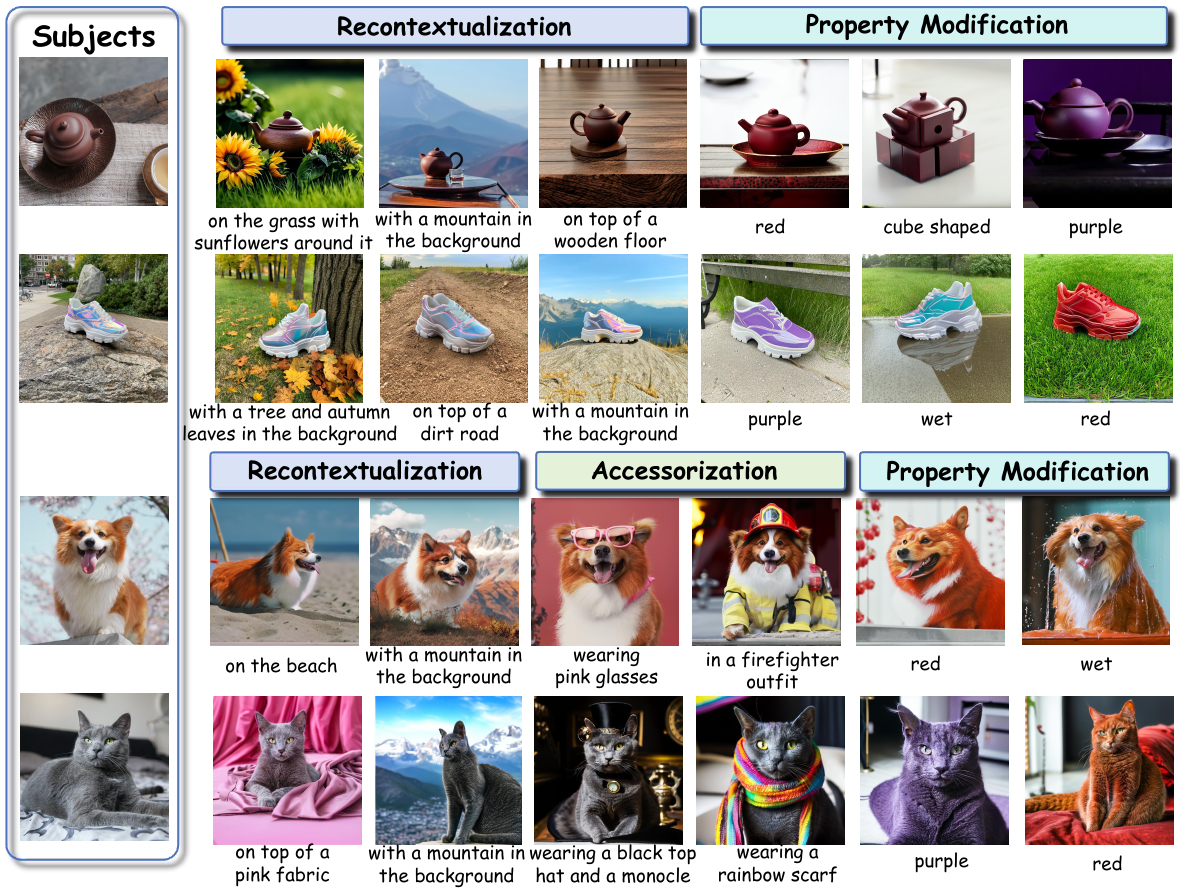}
    \caption{Qualitative results of subject-driven customization with DCoAR.}
    \label{Fig:qual_subject1}
\end{figure*}

\begin{figure*}
    \centering
    \includegraphics[width=0.6\linewidth]{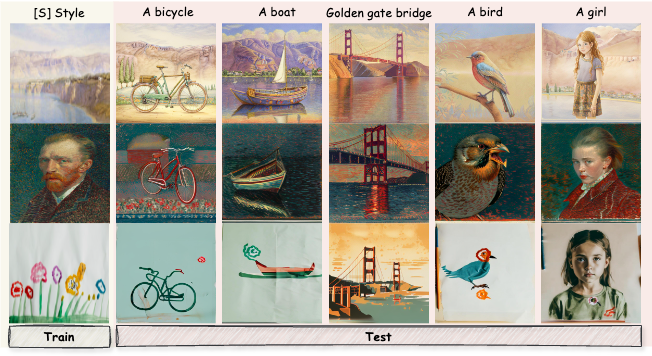}
    \caption{DCoAR demonstrates effective style personalization conditioned on only a single reference image.}
    \label{Fig:single_style}
\end{figure*}

\begin{figure*}
    \centering
    \includegraphics[width=0.99\linewidth]{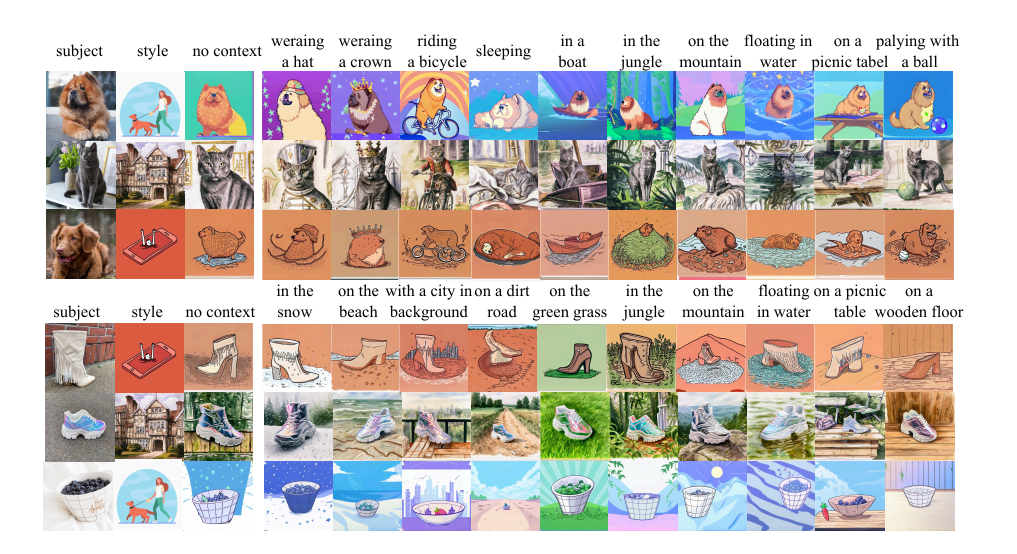}
    \caption{Impact of contextual prompts on image generation across subjects and styles.}
    \label{Fig:styledrop_recon}
\end{figure*}

\Cref{Fig:qual_subject1,Fig:qual_subject2,Fig:single_style,Fig:styledrop_recon} presents additional qualitative results of subject-driven personalization and style personalization.

\begin{figure}
    \centering
    \includegraphics[width=0.99\linewidth]{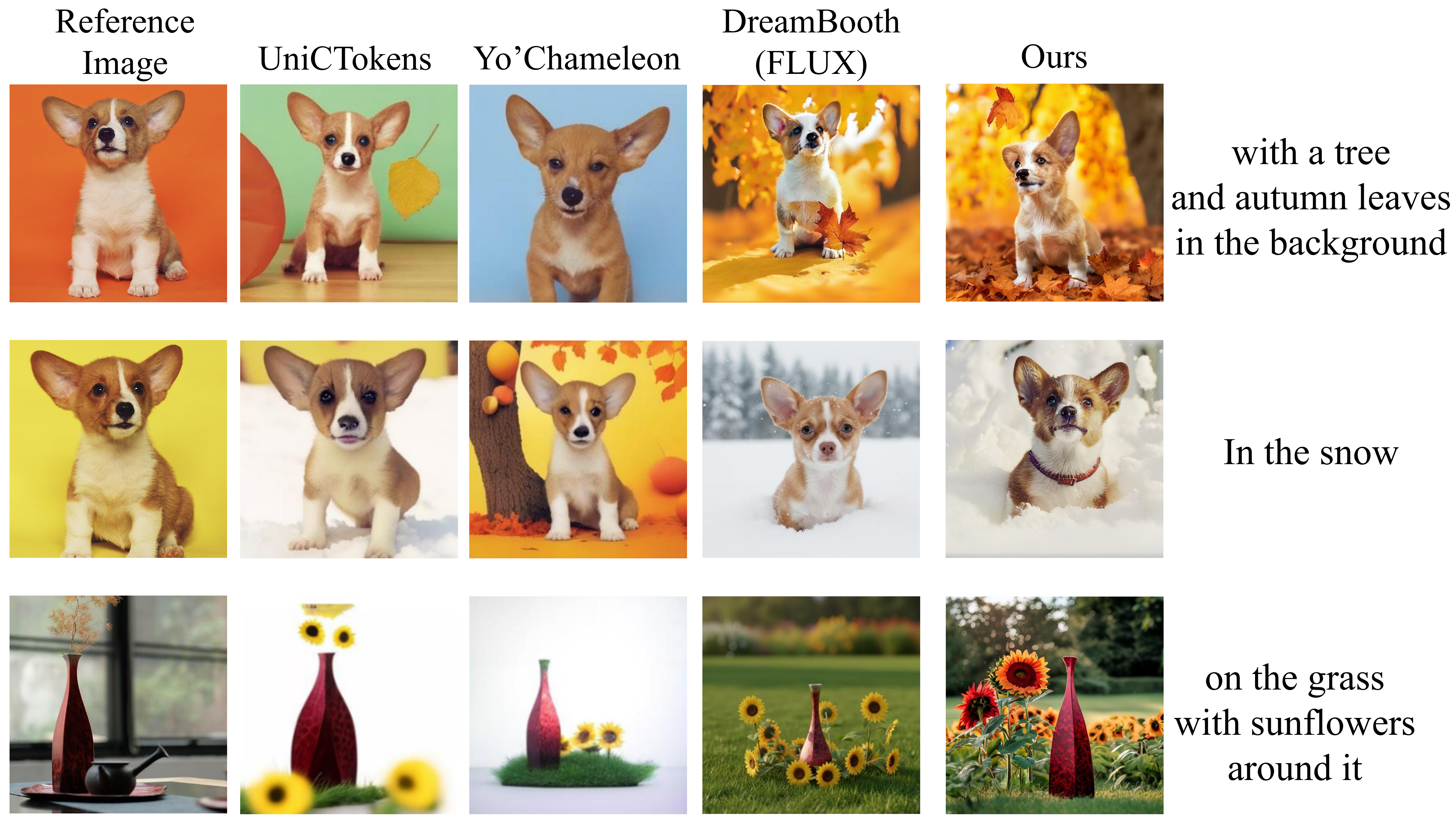}
    \caption{ \textbf{Additional qualitative comparison of subject-driven personalization}. We compare our DCoAR against two representative concept-injection methods (UniCTokens \cite{UniCTokens}, Yo’Chameleon \cite{yoChameleon}) and a state-of-the-art diffusion-based fine-tuning approach (DreamBooth \cite{ruiz2023dreambooth} with FLUX).}
    \label{Fig:comp—sup}
\end{figure}

We provide further visual evidence comparing DCoAR with competing baselines in \Cref{Fig:comp—sup}. Consistent with our main paper's findings, shallow concept-injection methods (UniCTokens\cite{UniCTokens}, Yo’Chameleon\cite{yoChameleon}) struggle to capture high-frequency details, often resulting in smoothed textures and lower visual fidelity. Conversely, while the adaptation-based method DreamBooth (utilizing the FLUX backbone) produces high-quality backgrounds, it exhibits significant semantic instability and identity drift; this is particularly evident in the second row, where the reference dog is erroneously rendered as a different breed. In contrast, DCoAR successfully preserves intricate subject details—such as specific fur patterns and material textures—while accurately adhering to complex background prompts, demonstrating the effectiveness of our Deep Concept Injection strategy.

\Cref{Fig:qual_subject1,Fig:qual_subject2} demonstrate that given only a few reference images, DCoAR successfully performs subject-driven generation across various tasks, including recontextualization (e.g., novel backgrounds), property modification (e.g., color and shape), and accessorization (e.g., adding glasses or outfits). The results demonstrate strong subject fidelity, adaptability to novel contexts, and fine-grained controllability.

\Cref{Fig:styledrop_recon} illustrates the effects of diverse contextual prompts on image generation for various subjects across different visual styles. Each row corresponds to a distinct subject, while each column varies the context.

\section{Limitations}

Despite the superior performance of DCoAR in personalized generation, we acknowledge the primary limitations that warrant future investigation:
\textbf{Sensitivity to Insertion Depth and Overfitting.}
Our framework relies on a carefully tuned Layer-wise Multimodal Context Learning (LMCL) strategy. As evidenced by our ablation studies, the model is sensitive to the depth of token injection. While shallow injection fails to capture identity details, \textbf{injecting tokens into overly deep layers} imposes excessive visual constraints. This leads to \textit{overfitting}, where the model becomes rigid and prioritizes pixel-level adherence to the reference image over the semantic control of the text prompt, hindering effective re-contextualization and stylization. Developing an adaptive mechanism to automatically determine the optimal insertion depth for different subjects remains an open problem.

\clearpage

\bibliographystyle{plain}
\bibliography{main}

\end{document}